\documentclass[letterpaper, 10 pt, conference]{ieeeconf}  
\usepackage{booktabs}
\IEEEoverridecommandlockouts                              

\usepackage{stfloats}

\usepackage{graphics} 
\usepackage{epsfig} 
\usepackage{mathptmx} 
\usepackage{times} 
\usepackage{amsmath} 
\usepackage{amssymb}  
\usepackage[T1]{fontenc}
\usepackage{cite}

\title{\LARGE \bf
Bio-inspired reconfigurable stereo vision for robotics using omnidirectional cameras
}

\author{Suchang Chen$^{1}$, Dongliang Fan$^{1}$, Junpeng Chen$^{1}$, Haichao Wang$^{1}$, Huijuan Feng$^{1}$, \\and Jian S Dai$^{1}$, \IEEEmembership{Fellow,~IEEE}
\thanks{*This work was supported by the Key Program of the National Natural Science Foundation of China (grant 52335003), the Postdoctoral Fellowship Program (Grade B) of China Postdoctoral Science Foundation (grant GZB20230286), the Science, Technology and Innovation Commission of Shenzhen Municipality (grant ZDSYS20220527171403009), and the High level of special funds from Southern University of Science and Technology, Shenzhen, China (grant G03034K003), China Postdoctoral Science Foundation (grant 2024M761294). (\textit {Corresponding authors: Dongliang Fan, fandl@sustech.edu.cn; Jian S Dai, jian.dai@kcl.ac.uk)}}
\thanks{$^{1}$Suchang Chen, Dongliang Fan, Junpeng Chen, Haichao Wang, Huijuan Feng and Jian S Dai are with the Shenzhen Key Laboratory of Intelligent Robotics and Flexible Manufacturing Systems, Southern University of Science and
Technology, Shenzhen 518055, China.
       }%
}

\begin{document}

\maketitle
\thispagestyle{empty}
\pagestyle{empty}

\begin{abstract}

This work introduces a novel bio-inspired reconfigurable stereo vision system for robotics, leveraging omnidirectional cameras and a novel algorithm to achieve flexible visual capabilities. Inspired by the adaptive vision of various species, our visual system addresses traditional stereo vision limitations, i.e., immutable camera alignment with narrow fields of view, by introducing a reconfigurable stereo vision system to robotics. Our key innovations include the reconfigurable stereo vision strategy that allows dynamic camera alignment, a robust depth measurement system utilizing a nonrectified geometrical method combined with a deep neural network for feature matching, and a geometrical compensation technique to enhance visual accuracy, even in very limited visual overlap (\(\sim\)56°). Implemented on a structural reconfigurable robot, this vision system demonstrates its great adaptability to various scenarios by switching its configurations of \(\sim\)207° monocular with \(\sim\)76° binocular field for fast target seeking and \(\sim\)95° monocular with \(\sim\)140° binocular field for detailed close inspection.

\end{abstract}

\section{INTRODUCTION}

Stereo vision techniques, enabling robots to perceive the world in three dimensions, have been pivotal in expanding the boundaries of robotic visual capabilities \cite{alfalahi_concentric_2020,van_duong_large-scale_2021}. From factory assembly lines to uncharted territories, stereo vision-equipped robots capable of accurate depth perception and enhanced spatial understanding undertake tasks with unprecedented exploration capacity and working precision \cite{yang_visual_2021,schuster_distributed_2019}. Binocular depth estimation is the primary method for obtaining spatial information in machine vision \cite{poggi_synergies_2021,flores-fuentes_3d_2023}. The common practice involves positioning two cameras at different viewpoints using pin-hole camera models for depth information acquisition. To improve measuring accuracy and reliability, the binocular camera is often enhanced with infrared structured light or line laser, employing homologous point matching methods, including epipolar alignment and feature extraction\cite{laga_survey_2022,papadimitriou_epipolar_1996}. Present models and products such as Intel's RealSense \cite{keselman_intelr_2017} and Microsoft's Kinect \cite{jungong_han_enhanced_2013} are frequently used as vision information sources for robots, which are widely applied in various robotic fields, such as mobile robots for navigation and field exploration \cite{placed_survey_2023}, and service robots for object recognition and manipulation \cite{cong_comprehensive_2023}. The hardware of existing stereo vision systems commonly consists of a pair of co-planar pin-hole cameras. However, this monotonous hardware with a limited field of view (FOV), \(\sim\)60-90° \cite{fu_application_2020}, confines the spatially visual performance in stereo vision systems for practical and versatile applicable scenarios.
   \begin{figure}[b]

      \centering
      \includegraphics[scale=0.4]{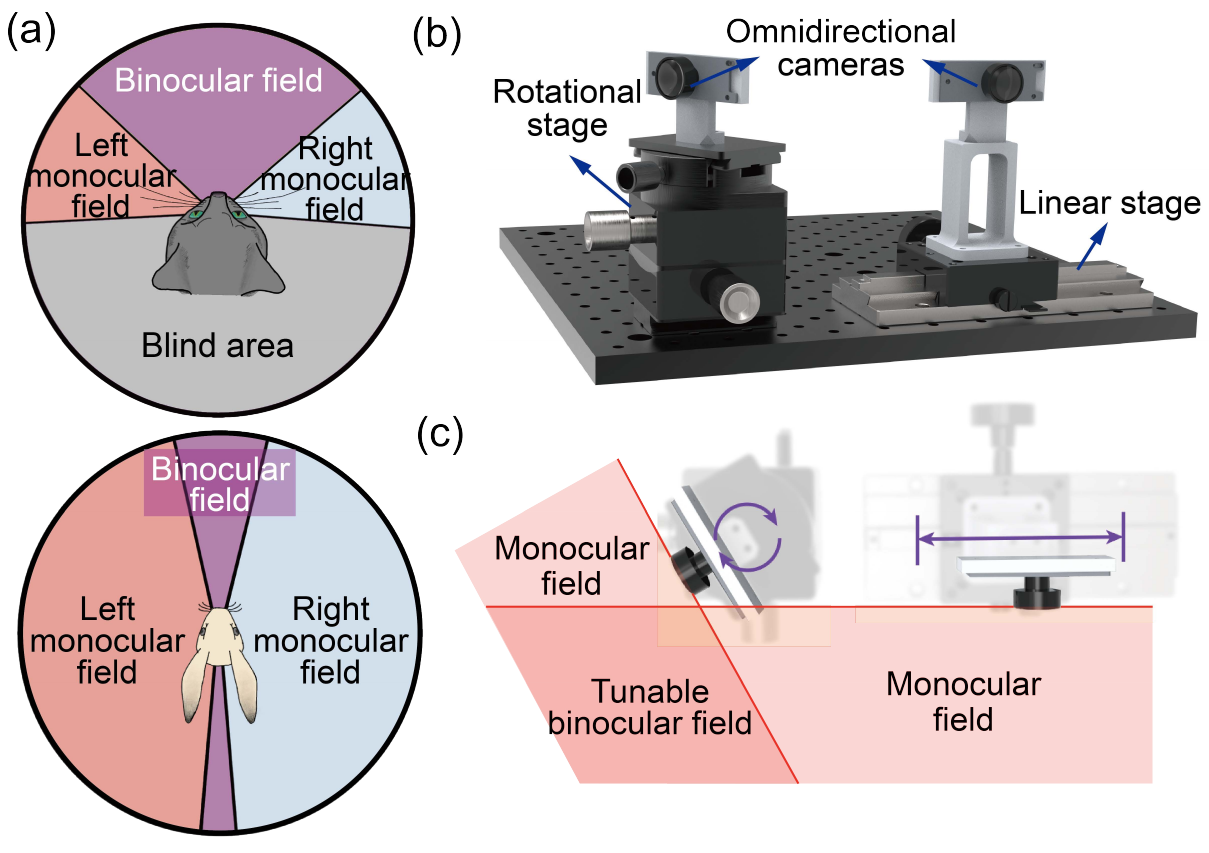}\\
      \caption{ Bio-inspired reconfigurable stereo vision system. (a) The visual field of the monocular and binocular field of cat and rabbit. (b) The physical platform of the visual system with two omnidirectional cameras. (c) The visual field of the reconfigurable vision.}
      \label{figurelabel}
      
   \end{figure}
   
To address the problems mentioned above, we turn to nature for heuristics. Present robots with bio-inspired designs have exhibited profound impacts in various intrinsic structures and biomimetic motions, including quadruped running, entangled grasping, and continuous jumping \cite{zhang_trot_2018, fan_self-shrinking_2022,aubin_powerful_2023}. Furthermore, the metamorphic design is inspired by the morphology variation and performance development during the growth process of typical species. Promising progress of the metamorphic design, such as various gaits of a quadruped robot realized by changing the relative position of the joint axis \cite{fu_stability_2023}, swimming and crawling modes switching by tuning the limbs of the turtle robot \cite{baines_multi-environment_2022}, and different moving strategies of the transformable origami exoskeleton \cite{miyashita_robotic_2017}, has been made to enhance their locomotion characteristics and adapt to versatile environments and tasks. Most of the existing bio-inspired designs focus on their mechanical structures and ignore the abundant vision features. Fig. 1(a) illustrated that Herbivores, e.g., rabbits, possess a panoramic FOV (360°) but narrow binocular field (\(\sim\) 30°)\cite{hughes_topographical_1971}. In contrast, carnivores and omnivores, like cats, possess a confined FOV (\(\sim \) 186°) but a broad binocular field (\( \sim \) 98°) \cite{hughes_supplement_1976}. The differences in FOV among different species are due to the alignment of the eyes. A reconfigurable stereo vision system would benefit robots by enabling them to explore different scenarios with tunable vision features.

To achieve the proposed vision system, a vision workflow algorithm that matches the system's requirements and features is necessary. Currently, common vision algorithms rely on the epipolar constraint and the pinhole camera assumption, while also presuming that the two cameras in a stereo vision system are fixed on the same plane, have closely matched focal lengths, and are horizontally aligned.\cite{laga_survey_2022,papadimitriou_epipolar_1996}. The current algorithmic limitation restricts the realignment of the cameras and confines the reconfigurability of the mainstream stereo vision systems. Furthermore, the tunable FOV envisaged with a reconfigurable physical platform also proposes a new challenge for the vision algorithm. 

Therefore, a biomimetic reconfigurable stereo vision system is proposed that can tune the alignment of the two omnidirectional cameras to create different FOV and binocular fields through a reconfigurable platform (Fig. 1(b-c)). The vision workflow algorithm modifies the nonrectified geometrical method \cite{campos_orb-slam3_2021}, which typically utilizes the retrieved optical path for depth measurement of the system to maintain the wide FOV. Additionally, a pre-trained deep neural network\cite{lindenberger_lightglue_2023} is employed to match local features reliably, and a geometrical compensation method is utilized to optimize and filter the results. This endows the system with the capability to collect stereo point clouds for multiple purposes, offering flexibility for a reconfigurable wide FOV vision system. Finally, the reconfigurable stereo vision system is deployed on a metamorphic robot, Origaker \cite{tang_origaker_2022}, providing this robot with various tunable vision features, including a large FOV for target-seeking and a large binocular field for detailed inspection, demonstrating the excellent adaptability for versatile applicable scenarios.

   \begin{figure*}[bp]
      \centering
      \includegraphics[scale=0.75]{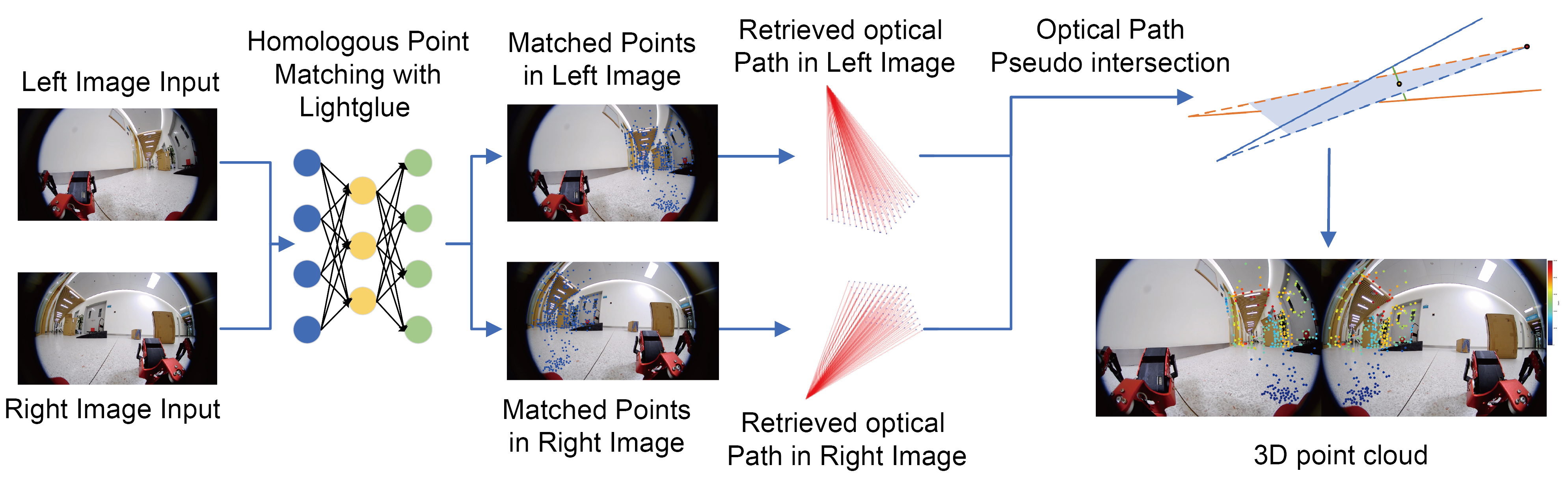}\\
      \caption{The vision pipeline of the proposed system. The pipeline performs mainly three tasks: Homologous point matching, optical path retrieval, and the computation of the pseudo intersection.}
      \label{figurelabel}
\end{figure*}

The main contributions of this article are as follows:

1. A reconfigurable stereo vision strategy was first proposed by tuning the alignment of the two installed omnidirectional cameras with a reconfigurable platform.

2. A depth measurement system for binocular vision was developed based on the improved nonrectified stereo method with a pre-trained deep neural network feature matcher and a geometrical compensation method, enabling precise depth measurement for a non-planar camera alignment.

3. The reconfigurable stereo vision system was deployed on a metamorphic robot to realize a tunable vision characteristic, demonstrating a large FOV for target seeking and a large binocular field for detailed inspection.

\section{Vision algorithm of the reconfigurable vision }

To obtain a biomimetic and wide FOV, two omnidirectional cameras are employed in our system. The conventional graphic process relies on methods like rectification, which typically results in a loss of both FOV and accuracy. Additionally, the technique is not applicable to reconfigurable design. Therefore, a geometric depth measurement approach \cite{campos_orb-slam3_2021} is utilized to maintain the edge region of omnidirectional images for depth measurement in non-planar camera settings. However, the native implementation of this method is constrained by precision limitations, with errors propagating from several aspects, such as extrinsic calibration and homologous point matching. While the point set acquired is feasible for navigation, it remains a challenge to use as a universal visual method. To address this problem, an off-the-shelf neural network module is employed to obtain reliable homologous points in complex and challenging environments robustly, and a geometric compensation method is integrated to optimize and filter the calculated spatial points. This approach enables the system to acquire a higher quality set of spatial point information, thereby enhancing its potential to be used as a general visual method. All spatial information obtained from our vision system is from this nonrectification method, including the setup of the system. Fig. 2 depicts the functional component of the vision pipeline.

\subsection{Optical Path Retrieve Model}

Stereo depth measurement follows a general pipeline: find homologous points in the overlapped field of view of binocular cameras, retrieve the incident optical path that connects the camera center and the real-world point, and then triangulate to find the spatial location. Retrieving the incident optical path is straightforward for commonly used pin-hole camera models since the optical path is not reflected. However, for omnidirectional cameras, special camera models are needed to map the optical path due to the use of a significant amount of barrel distortion to compress a wide scene into a limited image area. In our approach, the commonly adopted Kannala-Brandt model (Fig. 3) is utilized to address the distortion of the omnidirectional lens\cite{kannala_generic_2006}.

   \begin{figure}[b!]
      \centering
      \includegraphics[scale =0.25]{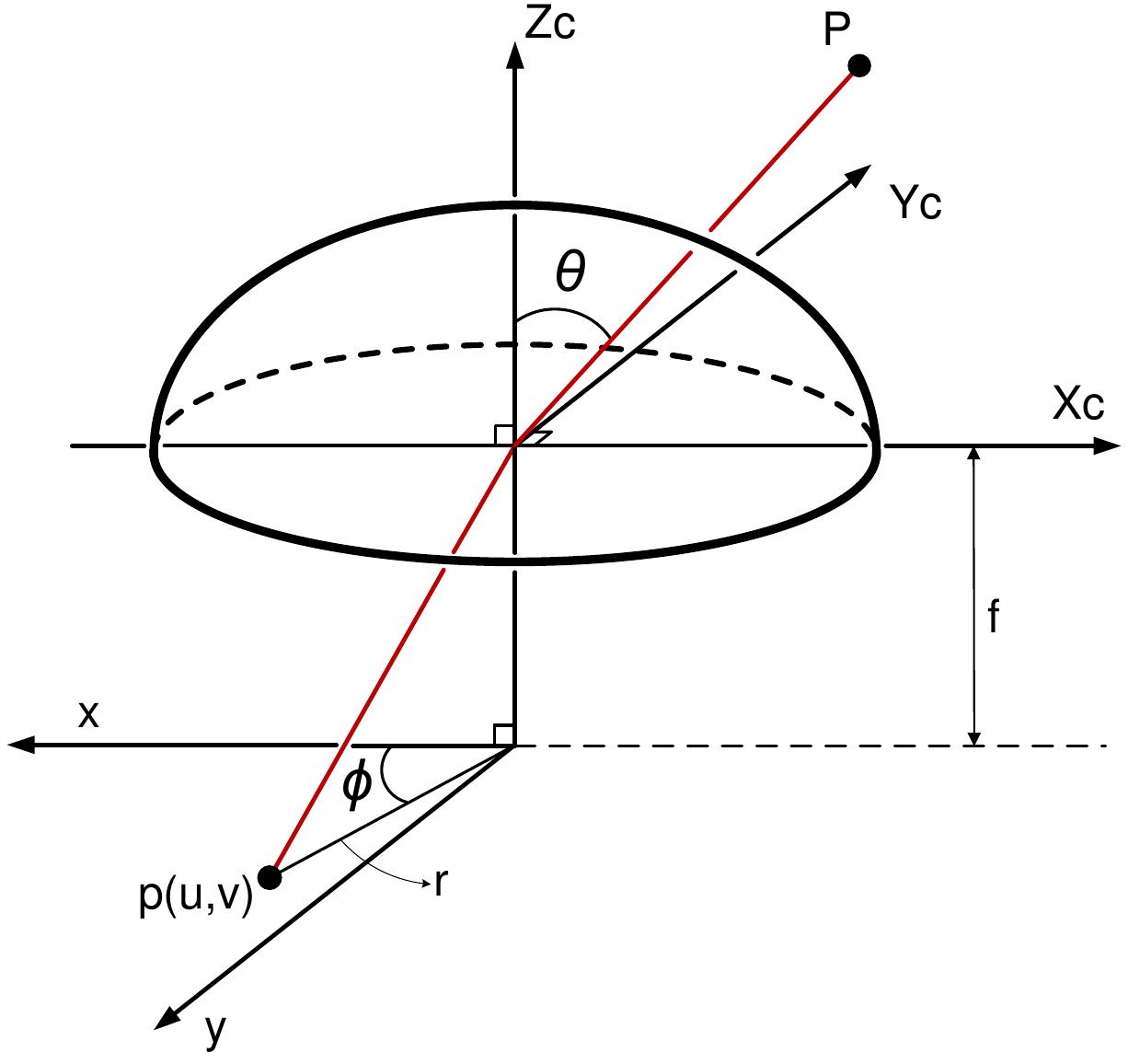}\\

      \caption{Schematic of the Kannala-Brandt camera model.}
      \label{figurelabel}
   \end{figure}

The image point of interest \(p(u, v)\) refers to the point where the incident ray passes spatial point \(P\) and projects onto the normalized image plane. \(f\) refers to the focal length, \(\theta\), the polar angle, refers to the angle between the principal axis and the incident ray, \(\phi\), the azimuthal angle, refers to the angle between the positive x-axis and the connection line of \(p(u, v)\) to the principal point. \(r\) is the distance of \(p(u, v)\) to the principal point. To identify this ray, two parameters, the polar angle \(\theta\) and the azimuthal angle \(\phi\), are needed. From the Kannala-Brandt camera model, \(\theta\) is known by: 

\begin{equation}
r = k_1 \theta + k_2 \theta^3 +k_3  \theta ^5+ k_4 \theta^7 + k_5 \theta^9 
\end{equation}
here, \(k_1\), \(k_2\), \(k_3\), \(k_4\) and \(k_5\) are the distortion parameters from camera calibration, 
\(r\) is computed with the image processing result and the intrinsic parameters. In this case, the equation could be used to solve \(\theta\) by numerically solving the polynomial.

The azimuthal angle \(\phi\) is then computed by:

\begin{equation}
\phi = {\rm arctan} \left ( \frac {v }{u } \right ) 
\end{equation}

With the knowledge that the incident optical path passes through the origin of the camera frame, the optical path of an image point of interest is now retrieved for the subsequent processes, including calibration and depth computation.

\subsection{Calibration}
To perform stereo vision depth measurements, both the intrinsic and extrinsic parameters of both cameras are needed. For intrinsic calibration, each of the two omnidirectional cameras is calibrated separately using the Kannala-Brandt model with OpenCV's built-in library \cite{kannala_generic_2006,zhang_flexible_2000}. For extrinsic calibration, the conventional approach typically involves rectifying the omnidirectional image before performing extrinsic calibration using the usual pin-hole camera model approach. However, the existing methods prove ineffective in our setup due to the non-planar camera placement and the narrow overlapped vision, which is often located at the edge regions in certain visual modes and would be cropped in the rectification. Hence, a novel method is developed for the extrinsic calibration of non-planarly placed omnidirectional cameras with possibly large camera viewpoint disparity in our reconfigurable systems.

Using the chessboard calibration board, OpenCV’s built-in grid point detector is leveraged to accurately determine the sub-pixel coordinates of the chessboard corners in the captured image. Each chessboard corner corresponds to a reverse incident optical path, which allows the system to identify a bundle of rays, each ray connecting the camera center with a grid point on the board. Utilizing the known distribution of the chessboard grid points, an optimization problem is formulated to identify a plane whose intersections with the ray bundle best fit the expected distribution. By solving this optimization problem, the 3D position of the chessboard corners could be precisely determined in the camera frame in one calibration picture.

Since our proposed method requires stereo vision overlap, the chessboard is positioned within the overlapping FOV of both cameras. This allows the system to determine the 3D spatial coordinates of the checkerboard grid points in each camera frame with the optimizing solution above. A least-squares fitting method\cite{arun_least-squares_1987} is then applied to fit the relative pose between the two camera frames, minimizing the reprojection error. Using this approach, the extrinsic parameters of the cameras are acquired.

\subsection{Homologous Point Matching}

In stereo vision systems, the identification of homologous points across different camera views is a critical step for accurate depth estimation and 3D reconstruction. However, despite recent advancements in stereo-matching techniques, conventional methods have proven inadequate for our flexible system, which requires robust homologous point matching in unrectified fisheye images with potentially extreme viewpoint variations. Traditional feature-based methods, such as SGBM and SIFT, are ill-suited for this task due to their inability to handle significant lens distortion and large angular separations between cameras, which are inherent in our setup. Similarly, state-of-the-art neural network-based stereo approaches, including DISPNET and RAFT-STEREO, have also fallen short. These methods are predominantly trained on datasets captured with pinhole cameras and small viewpoint disparities, and many of them output disparity maps rather than direct homologous point pairs, making them incompatible with our optical path methodology. To address these challenges, we employ \cite{lindenberger_lightglue_2023}, an advanced deep neural network architecture for local feature matching, leveraging its pre-trained weights to provide a robust and efficient solution for demanding conditions such as large viewpoint changes, limited visual overlap, and challenging lighting environments. Without additional training or fine-tuning, LightGlue enables accurate and reliable homologous point identification in complex scenarios, overcoming the limitations of traditional and neural network-based stereo-matching methods. This is particularly significant for applications involving fisheye cameras and large baseline configurations, where conventional techniques often fail to deliver satisfactory results. A comparison of common used SGBM algorithm and LightGlue's matching result is displayed in Fig. 4, where SGBM algorithm produces sparse matches with a notable fraction of errors. In contrast, LightGlue generates a denser set of matching pairs with significantly improved accuracy.

\subsection{Depth Computation Through Pseudo Intersection of Retrieved Optical Paths}
The spatial location of the homologous point is calculated by triangulation of the retrieved optical path and extrinsic parameters of the camera set. Ideally, these physical points lie at the intersection of the respective optical paths on the intersection plane (determined by the two camera frame origins and the ideal optical path intersection). However, in practice, due to matching errors and image degeneration throughout the hardware and workflow, it is mathematically challenging for two optical paths to intersect precisely. To address this issue, a geometric correction mechanism is introduced that projects the two optical paths onto a predicted intersection plane to find a pseudo intersection to represent the spatial location of the homologous point (Fig. 5). The predicted intersection plane is determined by the origins of the two camera frames and the midpoint of the shortest distance between the two optical paths\cite{han_nearest_2010}. \(S\) is the midpoint between the two optical paths, \(L_l\) and \(L_r\). The two projected optical paths, \(l_l\) and \(l_r\) intersect on the predicted plane to form the pseudo intersection \(S'\).

   \begin{figure}[bp!]
      \centering
      \includegraphics[scale=0.29]{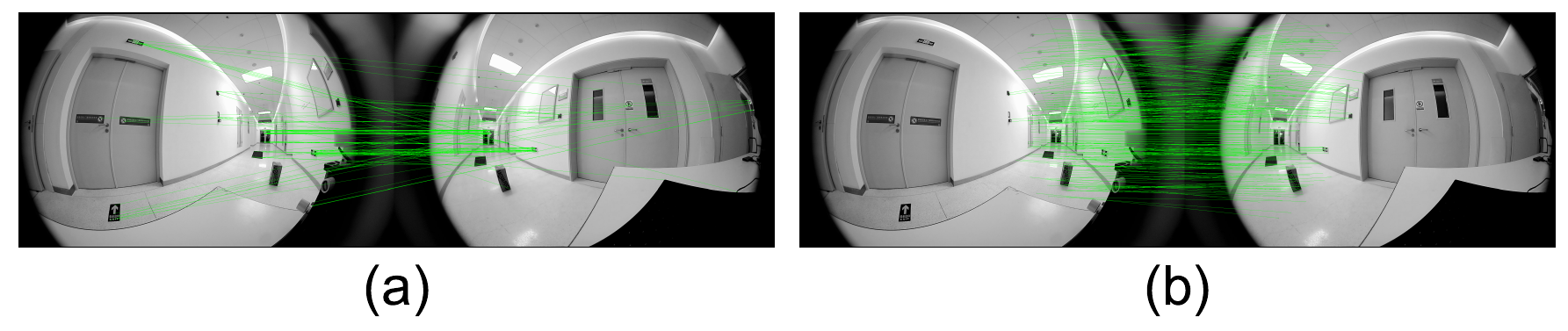}\\

      \caption{Stereo matching results of (a) SGBM algorithm and (b) pre-trained LightGlue.} 
      \label{figurelabel}
   \end{figure}

   \begin{figure}[tp!]
      \centering
      \includegraphics[scale=0.35]{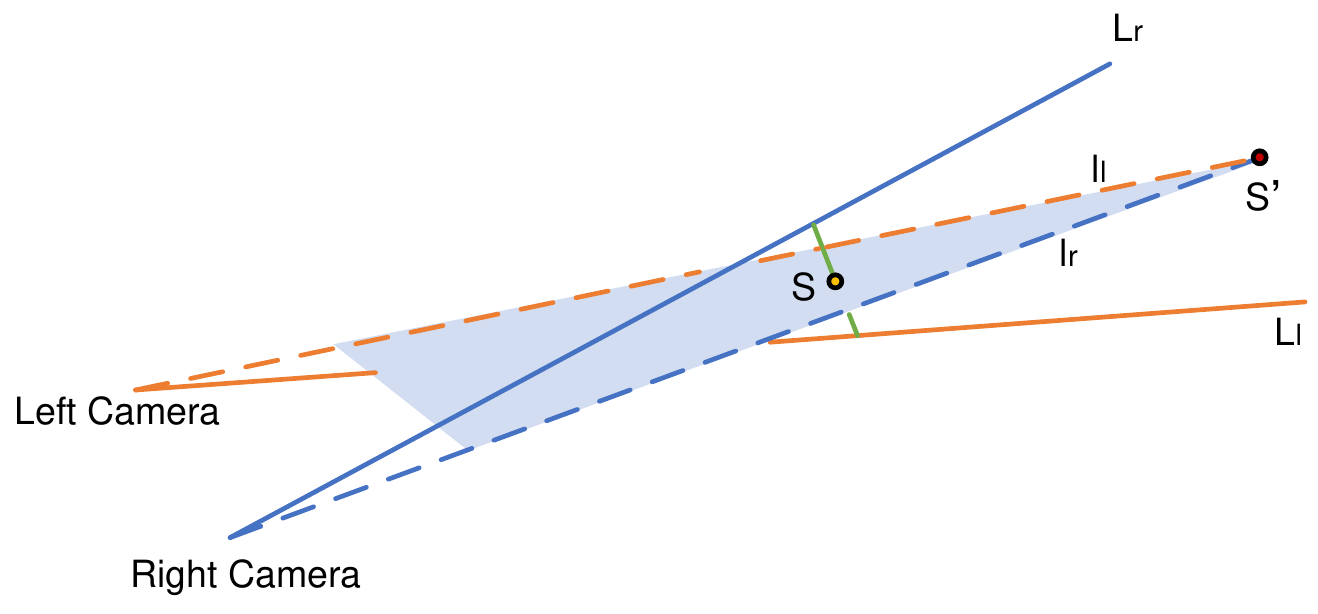}\\

      \caption{Schematic of the pseudo intersection formation.}
      \label{figurelabel}
   \end{figure}

This proposed pseudo intersection correction method, showing its great monotonicity against different matching errors, has presented its advantages in both error minimizing and error filtering. To minimize errors, finding the proposed pseudo intersection provides a more precise and robust guess for noisy or mismatched inputs. In the current workflow, the primary source of error arises from mismatched homologous points, which are common in the vision pipeline, especially in the presence of distortion and large camera angle separations. Vertical mismatches, often occurring due to the lack of feature points on vertical lines, are particularly frequent in practice. However, vertical mismatches remain a valuable visual resource for stereo vision applications, as real-world objects, which are typically upright and grounded, produce vertical lines in images. Unlike the original nonrectification method, which discards points with vertical mismatches, the proposed method retains this critical information, improving depth estimation accuracy.

\section{Experiments of the reconfigurable vision}

\subsection{Hardware of the Reconfigurable Vision}
The testing platform of the reconfigurable vision consists of a linear stage and a pinion translation stage to provide the relative translation and rotation of the two omnidirectional cameras (Fig. 6). The lenses of the cameras, SR1096A from SIR TEC, offer a wide angle of view with horizontal (H: 195.95°±4°), vertical (V: 159.8°±3°), and diagonal (D: 199.2°±4°) FOV. The optical sensor of the camera is the CV4002 CMOS Image Sensor by CVSENS. To fully utilize the large horizontal FOV of our reconfigurable vision system, the cameras were horizontally aligned during the tests.
   \begin{figure}[b!]
      \centering
      \includegraphics[scale = 0.3]{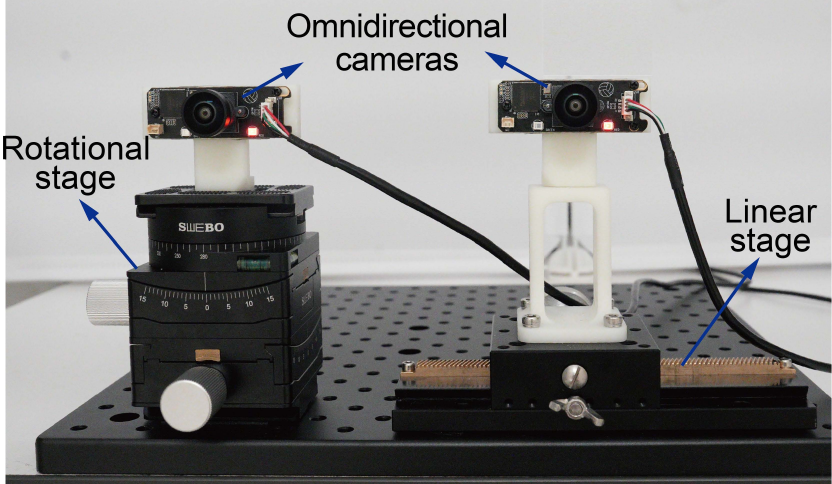}\\
      \caption{Hardware setup of the robot vision test platform with omnidirectional cameras and tunable stages.}
      \label{figurelabel}
   \end{figure}

To deploy the reconfigurable vision system, intrinsic parameters for both omnidirectional cameras were first calibrated. Subsequently, the system underwent a series of processes, including stereo extrinsic calibration, homologous point matching, and final depth computation. The results showed consistency with our intended design and biomimetic avatars.

\subsection{Characterization of FOV}

The reconfigurability of the system, with constant hardware, is demonstrated by varying the relative spatial positions of the two omnidirectional cameras. Due to the ultra-wide FOV of these cameras, translating them mainly adjusts the stereo system's baseline length, which in turn affects the accuracy of homologous point matching and depth computation. In contrast, rotating the cameras has a more profound impact on the characteristics of various vision modes, balancing the stereo measurement zone with the blind zone size, thus ensuring optimal performance in different biomimetic vision applications.

Moreover, to optimize depth measurement, the entire space is divided into three zones based on the relative poses of the two omnidirectional cameras:

\emph{1) Stereo Measuring Zone:} The region within the overlapped FOV used for accurate depth measurement.

\emph{2) Monocular Perception Zone:} Areas where only one camera has visibility.

\emph{3) Blind Zone:} Areas outside the FOV of both cameras, where visual information is not available.

This spatial division is crucial for effectively utilizing the stereo vision system and is primarily determined by the system configurations. Defining the field angle of each vision zone is also essential in evaluating the system's perceptual capabilities. A common approach is to determine the corresponding FOV width relative to different object plane positions. However, since the field angle of the vision zones in the proposed system may exceed 180°, this conventional method is not applicable. To address this limitation, we introduce the concept of the equivalent field angle to better define the system's FOV, as illustrated in Fig. 7 (a). 

This concept abstracts the binocular camera system as a single viewpoint located at the midpoint of the stereo baseline (assuming a symmetrical configuration). Given an equivalent object distance \( R \) for this viewpoint, the perception zones can be represented as arcs on a circle with a radius \( R \). The equivalent field angle of each zone is defined as the central angle corresponding to the arc. The three vision zones' equivalent field angle \(\theta_{stereo}\), \(\theta_{monocular}\), \(\theta_{blind}\) could be derived with known camera FOV \(\theta_{FOV}\), baseline length \(B\), camera main axis separation angle \(\theta_{separation}\) and object distance \(R\):

\begin{equation}
\left\{
\begin{aligned}
\theta_{stereo} &= u\left[ 1 - \frac{B}{2R}\cot\!\left(\frac{u}{2}\right) \right] \\
\theta_{blind} &= u + 2\,\arcsin\!\left[\frac{B}{2R}\,\sin\!\left(\frac{\pi+u}{2}\right)\right] \\
\theta_{monocular} &= 2\pi - \theta_{stereo} - \theta_{blind}
\end{aligned}
\right.
\end{equation}

\begin{figure*}[bp!]
      \centering
      \includegraphics[scale=0.75]{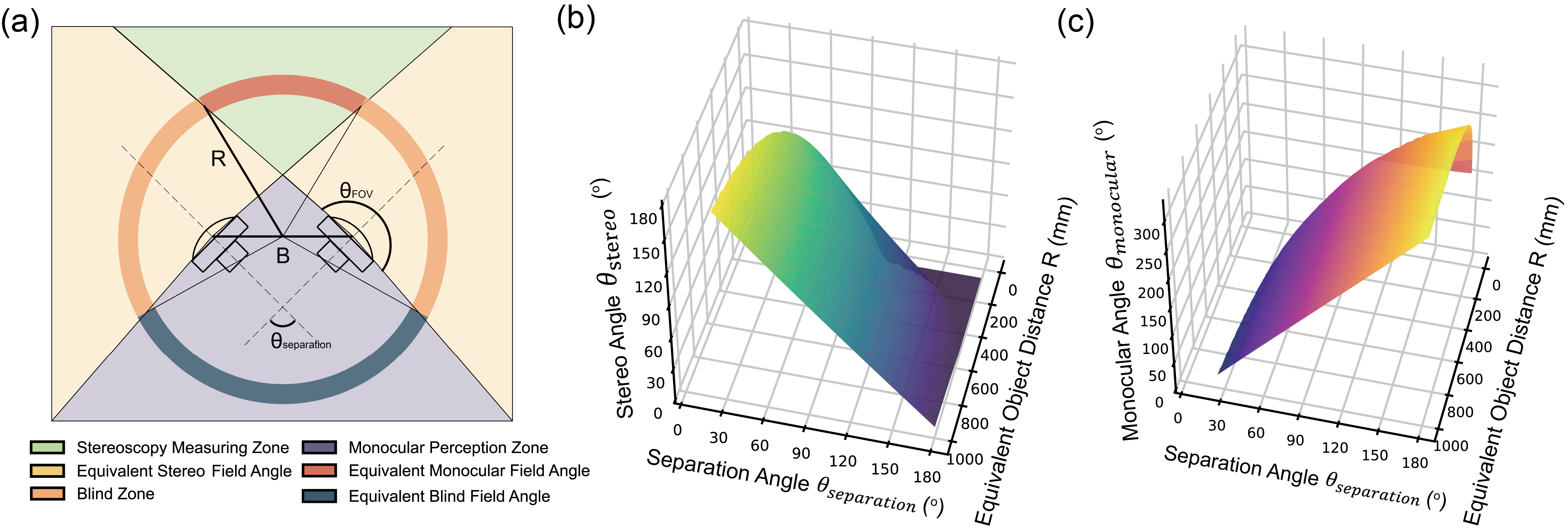}\\
      \caption{Equivalent field angle. (a) Perception zones and corresponding equivalent field angle. (b) Variation of stereoscopy measuring the zone's equivalent field angle \(\theta_{stereo}\) along with separation angle \(\theta_{separation}\) and equivalent object angle \(R\). (c) Variation of stereoscopy measuring the zone's equivalent field angle \(\theta_{monocular}\) along with separation angle \(\theta_{separation}\) and equivalent object angle \(R\).} 
      
      \label{figurelabel}
\end{figure*}

where
\begin{equation}
u = \theta_{FOV} - \theta_{separation}
\end{equation}

With the complex possible configurations of the system, Eq. (3) expresses the equivalent field angle of three perception zones in the general circumstance ($\theta_{separation} \geq \theta_{FOV} - \pi$, $R \geq \max\!\left(\frac{B}{2}\tan\!\left(\frac{u}{2}\right),\, \frac{B}{2}\right)$.) Fig. 7 (b) and (c) serves as a visualization for the variation of \(\theta_{stereo}\) and \(\theta_{monocular}\) with respect to separation angle \(\theta_{separation}\) and equivalent object angle \(R)\). With smaller \(\theta_{separation}\), the binocular camera tends to have more vision overlap, leading to a larger depth perception field, and vice versa.

\subsection{Characterization of Measurement Precision}

With the biomimetic and reconfigurable design strategy, the system can be deployed in any relative pose, provided that a common vision field exists. To evaluate the measurement precision of our system, two typical modes of the visual system and their unique characteristics are demonstrated using our vision test platform, as displayed in Fig. 8:
  \begin{figure}[tp!]
      \centering
      \includegraphics[scale = 0.48]{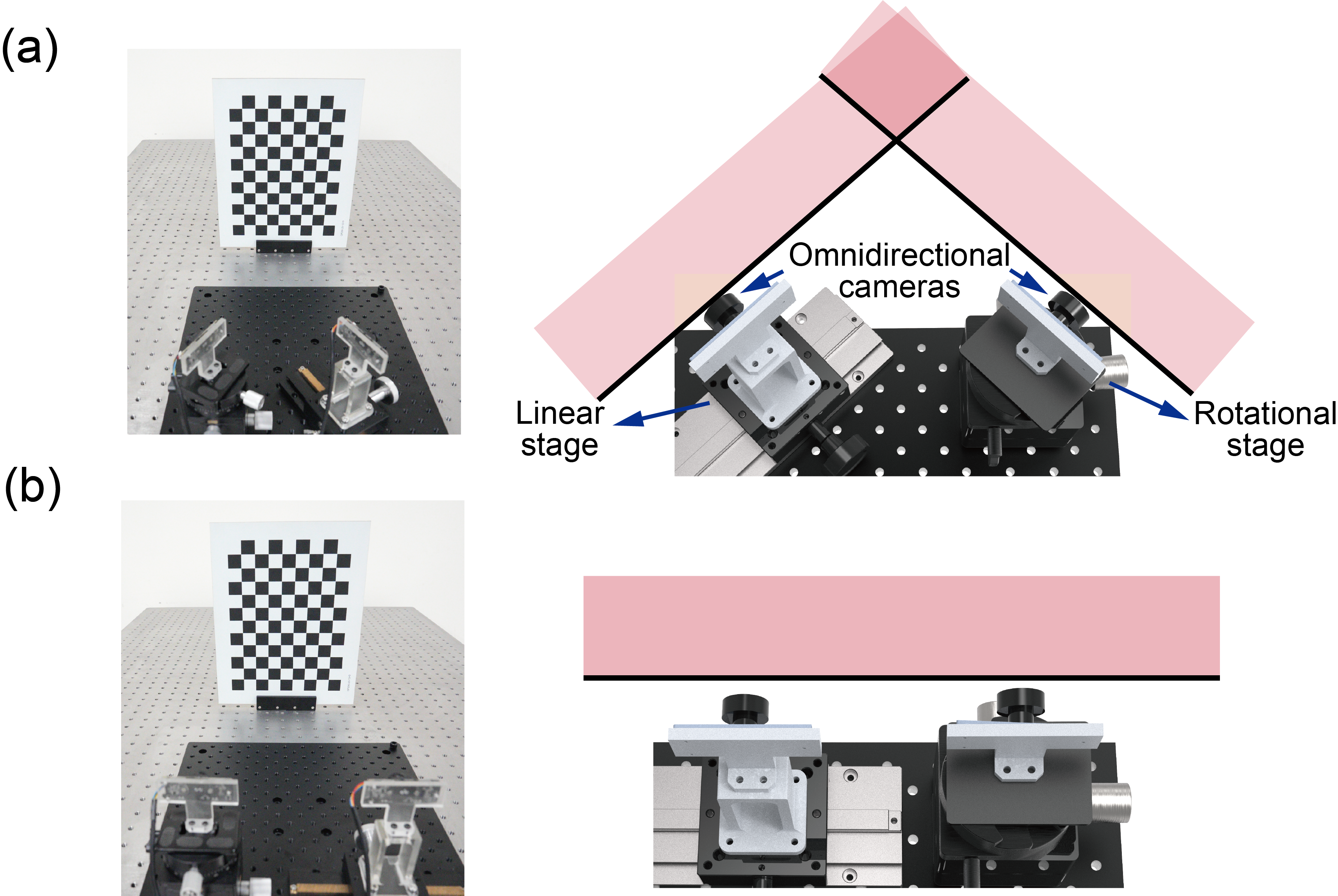}\\
      \caption{Reconfigurable vision precision testing experiment setup. (a) Herbivorous vision mode. (b) Human vision mode.}
      \label{figurelabel}
   \end{figure}

\emph{a) Herbivorous Vision mode: }This mode features an exceptionally large FOV (\(\sim\)265°) achieved by combining the view ranges of two omnidirectional cameras. It includes a relatively small angle of stereo vision for minimal 3D sensing (\(\sim\)56°). By testing on this configuration, the system’s ability for depth measurement on the lateral region of the omnidirectional camera is demonstrated.

\emph{b) Human Vision mode:} This mode includes comprehensive depth vision characterized by a broad depth field, with a \(\sim\)193° binocular vision field. Through testing on this configuration, the system’s ability to perform wide-range depth sensing is validated.

The field angle value presented here is calculated based on the actual experimental setup, with the equivalent object distance \( R \) set to 350 mm. A checkerboard target (20 mm \(\times\) 20 mm \(\times\) 350 mm), along with its holder, is mounted on an optical breadboard alongside the vision system, and its grid points are measured using different vision approaches. In human vision mode, where the conventional rectified extrinsic calibration method remains applicable due to the wide vision overlap, three measurement processes are conducted:

\emph{1) The conventional rectified binocular method:} The camera's extrinsic parameters are estimated through rectified calibration pictures, and the measured grid points are also found on rectified pictures. The spatial measurement assumes the epipolar constraint is satisfied during the rectification process and directly finds the disparity through the difference of \(x\) coordinates of the grid points. The depth \(Z\) is calculated by: 
 
 \begin{equation}
Z = \frac{f \cdot B}{d}
\end{equation}
where \(f\) represents the focal length, \(B\) represents the length of baseline, and \(d\) represents the pixel disparity. The 3D location is found by reprojecting the 2D points to 3D space.

\emph{2) The nonrectified optical path method:} With extrinsic parameters still estimated through a rectified calibration picture set, triangulation is performed using the retrieved optical path. The intersection is simply computed as the midpoint of the closest distance between the two incident optical paths.

\emph{3) The proposed geometrical-compensation nonrectified optical path method:} The proposed method uses the nonrectified calibration method introduced in Section $\uppercase\expandafter{\romannumeral 2}. B$ for extrinsic parameters and uses the pseudo intersection method introduced in Section $\uppercase\expandafter{\romannumeral 2}. D$ for the 3D measurement.

For the herbivorous vision mode, which contains a limited vision overlap (\(\sim\)56°), the conventional extrinsic calibration method fails because, after rectification, the checkerboard gets cropped in the process, preventing successful corner detection. Thus, only the measurement result of the proposed method is shown, as the other two methods cannot be performed without extrinsic parameters.

The depth measurement results of the two modes are shown in Fig. 9(a)–(d) in the following order: human mode conventional binocular method, human mode non-rectified method, human mode proposed method, and herbivorous mode proposed method. The absolute errors of depth measurement are 0.793 mm, 2.36 mm, 1.08 mm, and 1.44 mm, respectively. The relative errors are 0.227\%, 0.675\%, 0.308\%, and 0.413\%, respectively. The experimental results demonstrate the system's high measurement precision while preserving a wide FOV without rectification, even in extreme conditions such as narrow visual overlaps (\(\sim\)56°). The equivalent field angle of perception zones in all four test groups is shown in Fig. 9(e).

   \begin{figure}[bp!]
      \centering
      \includegraphics[scale=0.4]{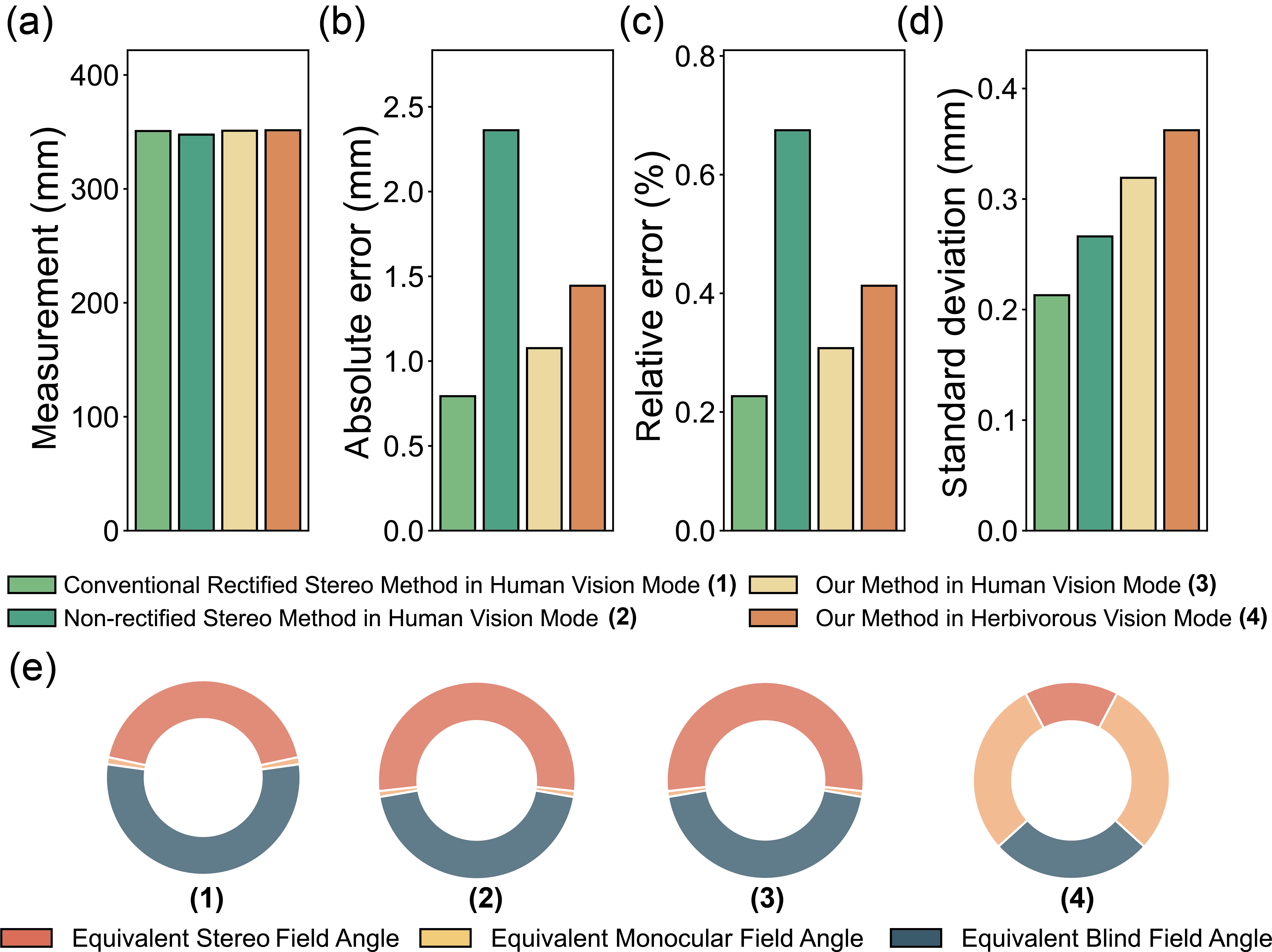}\\

      \caption{Depth Measurement value and error of grid point with different methods. (a) Measurement. (b) Absolute error. (c) Relative error. (d) Standard deviation. (e) Equivalent field angle profile of the three perception fields in all corresponding methods.}
      \label{figurelabel}
   \end{figure}

Therefore, our proposed method achieves measurement precision comparable to current baseline methods, even in extreme conditions where depth must be detected within the lateral regions of the omnidirectional image. The high precision further validates the system’s capacity to fully exploit its entire camera FOV for depth sensing, supporting the biomimetic principle that all overlapping vision fields can be used for stereo measurement, thereby justifying the proposed system’s extremely wide FOV. Meanwhile, with acceptable measurement precision, the generally wider FOV also demonstrated the improved measurement ability.

\section{Demonstration of the reconfigurable vision}

To further demonstrate the potential application of our reconfigurable vision system, we deployed it on a metamorphic quadruped robot, Origaker, which is a reconfigurable robot capable of altering its shape and mobility features by shifting its closed eight-bar linkage trunk\cite{tang_origaker_2022}. Two omnidirectional cameras are installed on the robot's trunk linkage. The relative pose relationship between the robot's geometric model and cameras is demonstrated in Fig. 10. Consequently, the relative position of the two omnidirectional cameras changes as the robot's form transforms, acquiring different vision features. The transitions between the visual stages are automatically executed upon receiving a human-issued control command,

   \begin{figure}[tp]
      \centering
      \includegraphics[scale=0.34]{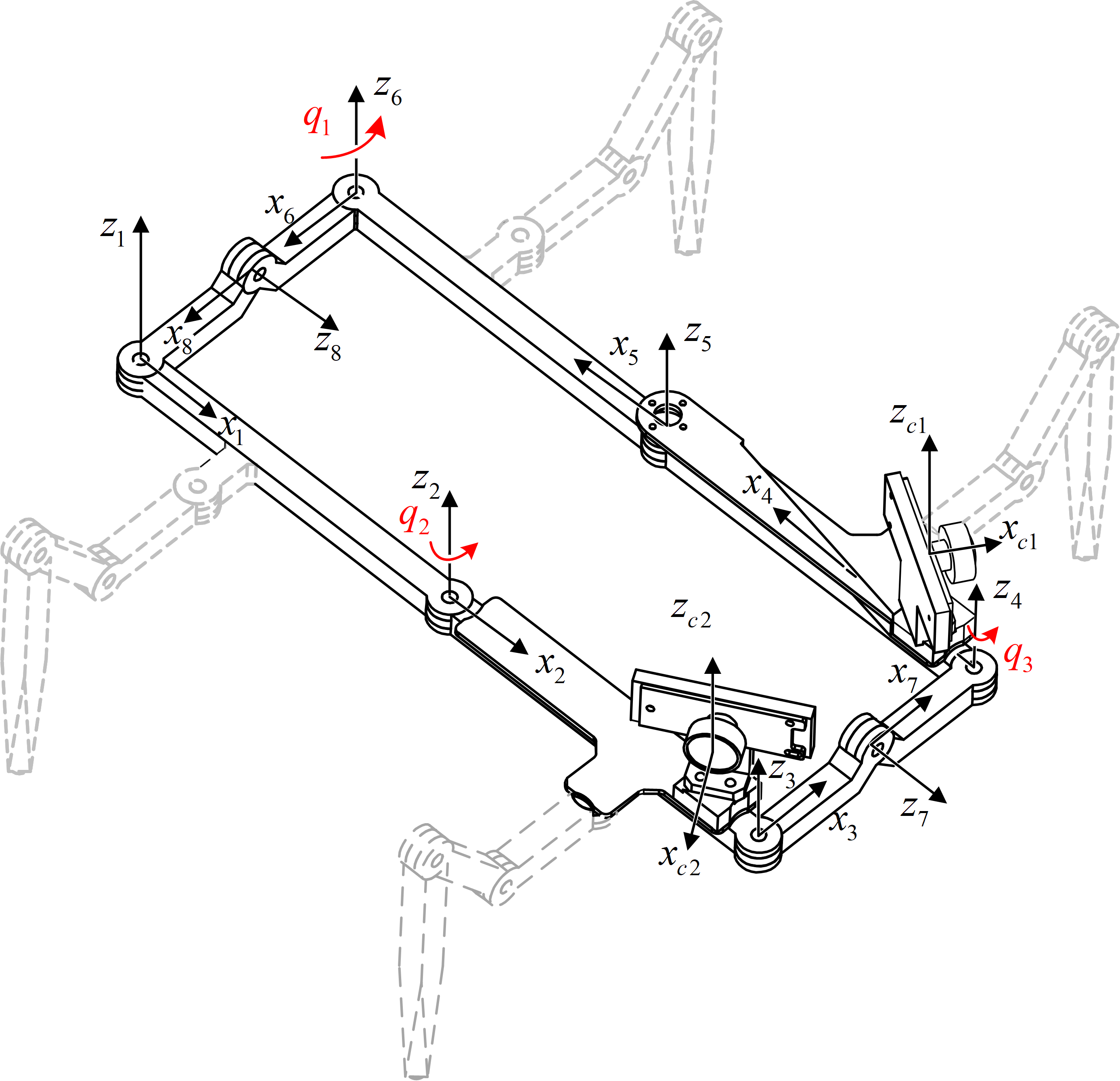}\\
      \caption{Sketch and coordinate system of the metamorphic trunk of Origaker with two installed cameras.}
      \label{figurelabel}
\end{figure}

   \begin{figure*}[bp!]
      \centering
      \includegraphics[scale=0.60]{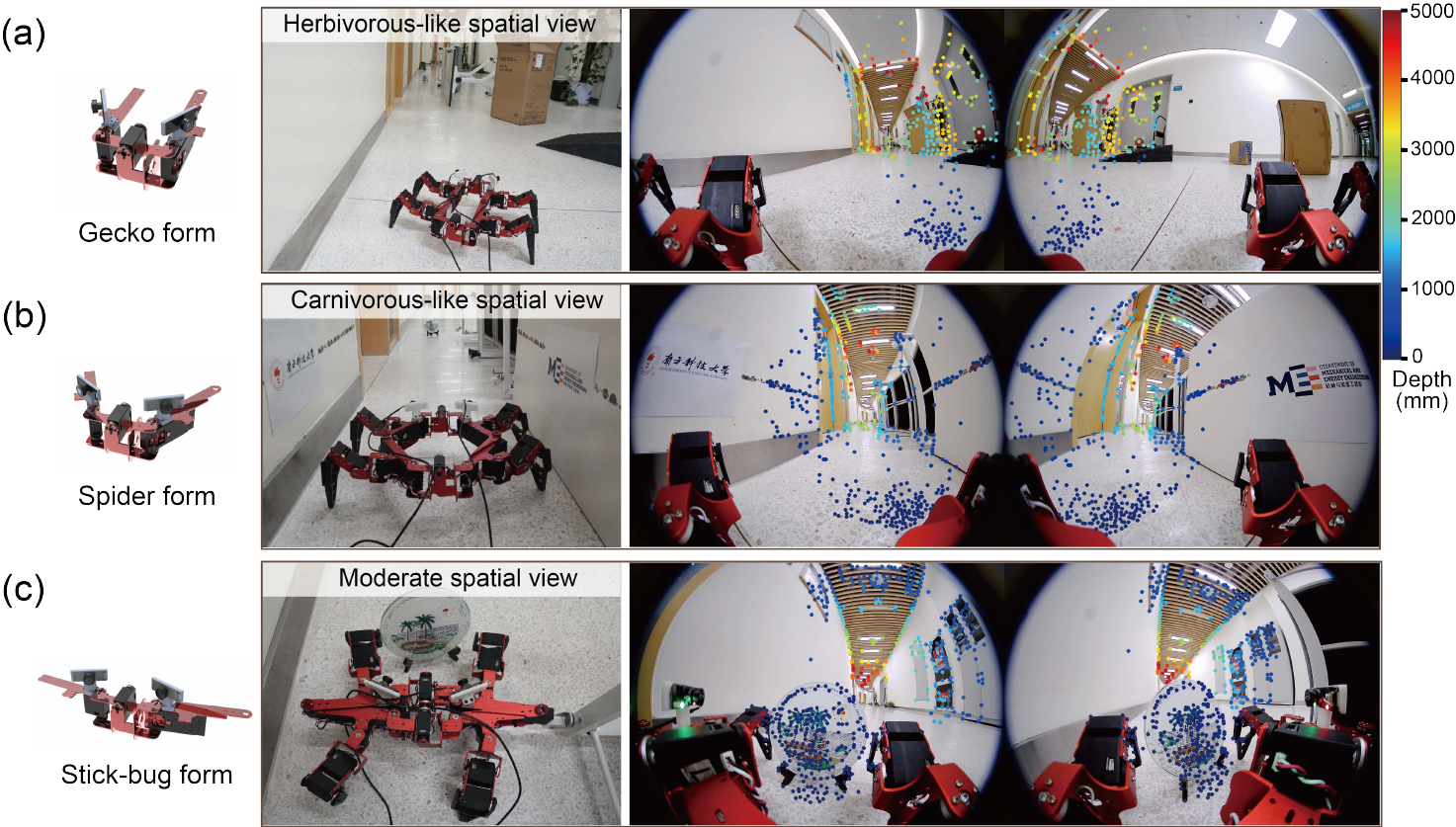}\\
      \caption{Demonstration of the proposed reconfigurable vision system on the metamorphic robot Origaker. Measured depth value colored according to the color bar.In the figure, the head link configuration of Origaker and the vision system is shown, along with the actual scenarios and corresponding vision information acquired from (a) The Gecko form. (b) The Spider form. (c) The Stick-bug form. }
      \label{figurelabel}
\end{figure*}

Origaker's torso employs a spatial closed-loop eight-bar linkage mechanism, exhibiting remarkable reconfigurability for mimicking structures of different creatures. This work focuses on three primary configurations: gecko, spider, and stick-bug modes. These configurations leverage the six-bar planar mechanism derived from the eight-bar spatial structure through kinematic degeneration (joints 7 and 8 are fixed). A modified Denavit-Hartenberg parameter method is adopted for robot modeling, where closed-loop equations are typically applied to solve joint angles for such mechanisms.

The closed-loop equation could be written as:
\begin{equation}
    T_{1}^2 T_2^{3} T_3^4 T_4^5 T_5^6 T_6^1= I
\end{equation}
where \( T_i^j \) represents the homogeneous transformation matrix from joint \( j \) to joint \( i \), ensuring the loop closure of the mechanism. This closed-loop equation establishes a system of nonlinear equations, which is solved using MATLAB's nonlinear least squares solver \texttt{lsqnonlin} under given \( q_i \) conditions to determine all joint angles. Subsequent rigid transformations yield camera poses $\{C_1\}$ and $\{C_2\}$ in our reconfigurable vision system, as shown in Fig. 10.

The gecko configuration provides a herbivorous-like spatial view, with a stereoscopic FOV of \( \sim \)76° and a monocular FOV of \( \sim \)207° (the widest one). The spider configuration provides a carnivorous-like spatial view, with a stereoscopic FOV of \( \sim \)136° and a monocular FOV of \( \sim \)87°. The stick-bug configuration provides a moderate spatial view, with a stereoscopic FOV of \( \sim \)140° (the widest one) and a monocular FOV of \( \sim \)95°. The FOV value mentioned here is calculated with actual robot setting parameters and an equivalent object distance \(R\) = 3000 mm. By adapting to these different configurations, the vision system on Origaker can exhibit varying perceptual characteristics, showcasing its versatility and robustness in different operational scenarios.

Here, to exhibit the adaptability of the reconfigurable vision, the Origaker was placed in a complex scenario to approach a specific object in a narrow corridor and inspect it in detail. The robot started in the gecko configuration, utilizing its wide FOV for extensive sensing and searching (Fig. 11(a)). With a relatively limited blind zone, the robot can obscurely avoid obstacles in almost all directions. The narrow field angle sensing in the center allows for minimal stereoscopic environment sensing to assist in-depth perception. After detecting the target object, the Origaker switched to the spider configuration to gain peripheral depth sensing (Fig. 11(b)). The blind zone is still less than 180°, ensuring a broad range of sensing capabilities. This configuration helps the robot accurately measure distances to the corridor walls and avoid collisions. Upon approaching the target object, the Origaker switched to the stick-bug configuration for detailed close-range observation (Fig. 11(c)). This mode optimizes camera alignment, facilitating the accurate matching of nearby homologous points, which other configurations struggle to achieve. However, the matching process becomes more challenging due to the significant difference in viewing angles, increasing the likelihood of occlusion, where distant objects are obstructed by closer ones. The stick-bug configuration, however, enhances depth measurement accuracy for nearby objects, offering a detailed perspective for precise shape reconstruction. Additionally, the transformation process and stereo vision features of each configuration are also demonstrated in Video 1.

\section{Conclusions and future work}
This research introduces a bio-inspired reconfigurable stereo vision system for robotics, using omnidirectional cameras to achieve adaptable visual features. Emulating animal visual adaptability, it overcomes the limitations of the traditional fixed alignment of cameras and narrow FOV. Key innovations include a reconfigurable stereo vision strategy, a robust depth measurement system using a modified nonrectified geometrical method and deep neural network, and the successful deployment of the metamorphic robot, Origaker. The system offers flexibility with different vision modes, such as a wide field for target seeking and a broad binocular field for detailed inspection, highlighting the potential of bio-inspired designs to enhance robotic vision adaptability in various scenarios. Future research will focus on refining the mechanical design to closely mimic eye alignment mechanisms, enabling a more comprehensive replication of the diverse visual strategies observed in different species. Additionally, having the system's transformation further automated with the task requirement is also a promising direction.


\begin{thebibliography}{99}

\bibitem{alfalahi_concentric_2020} H. Alfalahi, F. Renda, and C. Stefanini, “Concentric Tube Robots for Minimally Invasive Surgery: Current Applications and Future Opportunities,” IEEE Trans. Med. Robot. Bionics, vol. 2, no. 3, pp. 410–424, Aug. 2020. 
\bibitem{van_duong_large-scale_2021} L. Van Duong and V. A. Ho, “Large-Scale Vision-Based Tactile Sensing for Robot Links: Design, Modeling, and Evaluation,” IEEE Trans. Robot., vol. 37, no. 2, pp. 390–403, Apr. 2021.
\bibitem{yang_visual_2021} J. Yang, C. Wang, B. Jiang, H. Song, and Q. Meng, “Visual Perception Enabled Industry Intelligence: State of the Art, Challenges and Prospects,” IEEE Trans. Ind. Inf., vol. 17, no. 3, pp. 2204–2219, Mar. 2021.
\bibitem{schuster_distributed_2019} M. J. Schuster, K. Schmid, C. Brand, and M. Beetz, “Distributed stereo vision-based 6D localization and mapping for multi-robot teams,” Journal of Field Robotics, vol. 36, no. 2, pp. 305–332, Mar. 2019.
\bibitem{poggi_synergies_2021} M. Poggi, F. Tosi, K. Batsos, P. Mordohai, and S. Mattoccia, “On the Synergies between Machine Learning and Binocular Stereo for Depth Estimation from Images: a Survey,” IEEE Trans. Pattern Anal. Mach. Intell., pp. 1–1, 2021.
\bibitem{flores-fuentes_3d_2023} W. Flores-Fuentes, G. Trujillo-Hernández, I. Y. Alba-Corpus, J. C. Rodríguez-Quiñonez, J. E. Mirada-Vega, D. Hernández-Balbuena, F. N. Murrieta-Rico, and O. Sergiyenko, “3D spatial measurement for model reconstruction: A review,” Measurement, vol. 207, p. 112321, Feb. 2023.
\bibitem{laga_survey_2022} H. Laga, L. V. Jospin, F. Boussaid, and M. Bennamoun, “A Survey on Deep Learning Techniques for Stereo-Based Depth Estimation,” IEEE Trans. Pattern Anal. Mach. Intell., vol. 44, no. 4, pp. 1738–1764, Apr. 2022.
\bibitem{papadimitriou_epipolar_1996} D. Papadimitriou and T. Dennis, “Epipolar line estimation and rectification for stereo image pairs,” IEEE Trans. on Image Process., vol. 5, no. 4, pp. 672–676, Apr. 1996.
\bibitem{keselman_intelr_2017} L. Keselman, J. I. Woodfill, A. Grunnet-Jepsen, and A. Bhowmik, “Intel(R) RealSense(TM) Stereoscopic Depth Cameras,” in 2017 IEEE Conference on Computer Vision and Pattern Recognition Workshops (CVPRW). Honolulu, HI, USA: IEEE, Jul. 2017, pp. 1267–1276.
\bibitem{jungong_han_enhanced_2013} Jungong Han, Ling Shao, Dong Xu, and J. Shotton, “Enhanced Computer Vision With Microsoft Kinect Sensor: A Review,” IEEE Trans. Cybern., vol. 43, no. 5, pp. 1318–1334, Oct. 2013. 
\bibitem{placed_survey_2023} J. A. Placed, J. Strader, H. Carrillo, N. Atanasov, V. Indelman, L. Carlone, and J. A. Castellanos, “A Survey on Active Simultaneous Localization and Mapping: State of the Art and New Frontiers,” IEEE Trans. Robot., vol. 39, no. 3, pp. 1686–1705, Jun. 2023.
\bibitem{cong_comprehensive_2023} Y. Cong, R. Chen, B. Ma, H. Liu, D. Hou, and C. Yang, “A Comprehensive Study of 3-D Vision-Based Robot Manipulation,” IEEE Trans. Cybern., vol. 53, no. 3, pp. 1682–1698, Mar. 2023.
\bibitem{fu_application_2020} L. Fu, F. Gao, J. Wu, R. Li, M. Karkee, and Q. Zhang, “Application of consumer RGB-D cameras for fruit detection and localization in field: A critical review,” Computers and Electronics in Agriculture, vol. 177, p. 105687, Oct. 2020.
\bibitem{zhang_trot_2018} C. Zhang and J. Dai, “Trot Gait with Twisting Trunk of a Metamorphic Quadruped Robot,” J Bionic Eng, vol. 15, no. 6, pp. 971–981, Nov. 2018. 
\bibitem{fan_self-shrinking_2022} D. Fan, X. Yuan, W. Wu, R. Zhu, X. Yang, Y. Liao, Y. Ma, C. Xiao, C. Chen, C. Liu, H. Wang, and P. Qin, “Self-shrinking soft demoulding for complex high-aspect-ratio microchannels,” Nat Commun, vol. 13, no. 1, p. 5083, Aug. 2022.
\bibitem{aubin_powerful_2023} C. A. Aubin, R. H. Heisser, O. Peretz, J. Timko, J. Lo, E. F. Helbling, S. Sobhani, A. D. Gat, and R. F. Shepherd, “Powerful, soft combustion actuators for insect-scale robots,” Science, vol. 381, no. 6663, pp. 1212–1217, Sep. 2023.
\bibitem{fu_stability_2023} J. Fu, J. Chen, Z. Tang, Z. Wei, and J. S. Dai, “Stability Margin Based Gait Design on Slopes for a Novel Reconfigurable Quadruped Robot with a Foldable Trunk,” in 2023 IEEE International Conference on Robotics and Biomimetics (ROBIO). Koh Samui, Thailand: IEEE, Dec. 2023, pp. 1–7.
\bibitem{baines_multi-environment_2022} R. Baines, S. K. Patiballa, J. Booth, L. Ramirez, T. Sipple, A. Garcia, F. Fish, and R. Kramer-Bottiglio, “Multi-environment robotic transitions through adaptive morphogenesis,” Nature, vol. 610, no. 7931, pp. 283–289, Oct. 2022.
\bibitem{miyashita_robotic_2017} S. Miyashita, S. Guitron, S. Li, and D. Rus, “Robotic metamorphosis by origami exoskeletons,” Sci. Robot., vol. 2, no. 10, p. eaao4369, Sep. 2017.
\bibitem{hughes_topographical_1971} A. Hughes, “Topographical relationships between the anatomy and physiology of the rabbit visual system,” Doc Ophthalmol, vol. 30, no. 1, pp. 33–159, Sep. 1971.
\bibitem{hughes_supplement_1976} A. Hughes, “A supplement to the cat schematic eye,” Vision Research,
vol. 16, no. 2, pp. 149–IN2, Jan. 1976.
\bibitem{campos_orb-slam3_2021} C. Campos, R. Elvira, J. J. G. Rodriguez, J. M. M. Montiel, and J. D. Tardos, “ORB-SLAM3: An Accurate Open-Source Library for Visual, Visual–Inertial, and Multimap SLAM,” IEEE Trans. Robot., vol. 37, no. 6, pp. 1874–1890, Dec. 2021. 
\bibitem{lindenberger_lightglue_2023} P. Lindenberger, P.-E. Sarlin, and M. Pollefeys, “LightGlue: Local
Feature Matching at Light Speed,” in 2023 IEEE/CVF International Conference on Computer Vision (ICCV). Paris, France: IEEE, Oct. 2023, pp. 17 581–17 592.
\bibitem{tang_origaker_2022} Z. Tang, K. Wang, E. Spyrakos-Papastavridis, and J. S. Dai, “Origaker: A Novel Multi-Mimicry Quadruped Robot Based on a Metamorphic Mechanism,” Journal of Mechanisms and Robotics, vol. 14, no. 6, p. 060907, Dec. 2022.
\bibitem{kannala_generic_2006} J. Kannala and S. Brandt, “A generic camera model and calibration method for conventional, wide-angle, and fish-eye lenses,” IEEE Trans. Pattern Anal. Mach. Intell., vol. 28, no. 8, pp. 1335–1340, Aug. 2006.
\bibitem{zhang_flexible_2000} Z. Zhang, “A flexible new technique for camera calibration,” IEEE Trans. Pattern Anal. Machine Intell., vol. 22, no. 11, pp. 1330–1334, Nov. 2000. 
\bibitem{arun_least-squares_1987} K. S. Arun, T. S. Huang, and S. D. Blostein, “Least-Squares Fitting of Two 3-D Point Sets,” IEEE Trans. Pattern Anal. Mach. Intell., vol. PAMI-9, no. 5, pp. 698–700, Sep. 1987.
\bibitem{han_nearest_2010} L. Han and J. C. Bancroft, “Nearest approaches to multiple lines in n-dimensional space,” Crewes Res. Rep, vol. 22, pp. 1–17, 2010.


\end{thebibliography}
\end{document}